\documentclass[conference]{IEEEtran}
\IEEEoverridecommandlockouts
\usepackage{cite}
\usepackage{amsmath,amssymb,amsfonts}
\usepackage{graphicx}
\usepackage{textcomp}
\usepackage{xcolor}

\usepackage[hidelinks]{hyperref}

\usepackage{multirow}
\usepackage{booktabs}
\usepackage{amsmath}
\usepackage{algorithm}
\usepackage{algpseudocode}

\usepackage{color, colortbl}
\definecolor{Gray}{gray}{0.9}

\def\BibTeX{{\rm B\kern-.05em{\sc i\kern-.025em b}\kern-.08em
    T\kern-.1667em\lower.7ex\hbox{E}\kern-.125emX}}
\begin{document}

\title{EMIT - Event-Based Masked Auto Encoding for Irregular Time Series}

\author{
    Hrishikesh Patel$^{1}$, Ruihong Qiu$^{1}$, Adam Irwin$^{1,2}$, Shazia Sadiq$^{1}$, Sen Wang$^{1}$ \\
    $^1$The University of Queensland, Australia \\
    $^2$Queensland Children’s Hospital, Australia \\
    \{hrishikeshm.patel, r.qiu, a.irwin, sen.wang\}@uq.edu.au, shazia@eecs.uq.edu.au \\
    \
    }

\maketitle

\begin{abstract}
Irregular time series, where data points are recorded at uneven intervals, are prevalent in healthcare settings, such as emergency wards where vital signs and laboratory results are captured at varying times. This variability, which reflects critical fluctuations in patient health, is essential for informed clinical decision-making. Existing self-supervised learning research on irregular time series often relies on generic pretext tasks like forecasting, which may not fully utilise the signal provided by irregular time series. There is a significant need for specialised pretext tasks designed for the characteristics of irregular time series to enhance model performance and robustness, especially in scenarios with limited data availability. This paper proposes a novel pretraining framework, EMIT, an event-based masking for irregular time series. EMIT focuses on masking-based reconstruction in the latent space, selecting masking points based on the rate of change in the data. This method preserves the natural variability and timing of measurements while enhancing the model’s ability to process irregular intervals without losing essential information. Extensive experiments on the MIMIC-III and PhysioNet Challenge datasets demonstrate the superior performance of our event-based masking strategy. The code has been released at \href{https://github.com/hrishi-ds/EMIT}{https://github.com/hrishi-ds/EMIT}.
\end{abstract}


\section{Introduction}
\begin{figure}[!t]
  \centering  \includegraphics[width=0.8\linewidth]{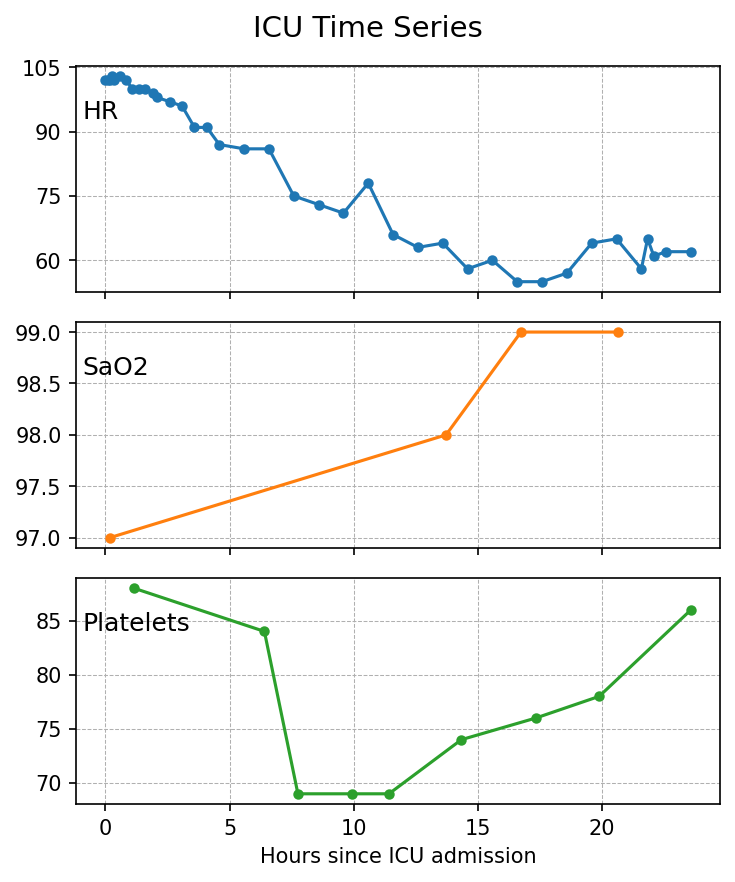}
  \caption{The plots display the measurements of heart rate, oxygen saturation, and platelet count for a patient in the ICU during the first 24 hours after admission. The data points are recorded at inconsistent intervals and the data collection is asynchronous among the three clinical variables.
  }
  \label{fig:irts}
\end{figure}

Irregular time series data, recorded at non-uniform intervals, are common in fields like finance, Internet-of-things, and healthcare. For instance, electronic health records, especially in intensive care units (ICUs) and emergency wards, often have irregular data (Fig.~\ref{fig:irts}). This irregularity often reflects the clinical decision-making and hence serves as a useful signal in time series modelling. Such data can be helpful for clinical prediction tasks like in-hospital mortality prediction, decompensation, and disease risk prediction~\cite{harutyunyan2019multitask, chenDynamicIllnessSeverity2018, peng2019temporal}.

Existing research in irregular time series modeling falls into two main categories: developing specialized architectures to handle irregularity and creating effective pretext tasks for self-supervised learning. Notable architectures include adaptations of Recurrent Neural Networks (RNNs), models based on ordinary differential equations, Transformer-based architectures, and graph-based methods. Additionally, to make use of limited annotated data, researchers have explored various self-supervised learning strategies, such as window-based forecasting, time-aware contrastive learning, and density-aware mask-based reconstruction. These methods highlight that irregularity is an essential component and serves as a useful signal in time series modeling.

Despite progress, working with irregular time series data remains challenging~\cite{shukla2021deep}. Firstly, in multivariate time series, observations often lack temporal alignment as they are recorded at different times. For example, lab results and ICU sensor data are not always recorded simultaneously. This asynchrony makes it difficult for models to learn patterns and capture temporal changes. Additionally, the density of data collection can vary, resulting in significant amounts of missing data. Secondly, even when features are recorded simultaneously, uneven gaps between observations can hinder the identification of useful patterns. Moreover, the scarcity of labelled data points further hampers the training of reliable models. In clinical settings, the high cost of labeling data exacerbates these difficulties. Therefore, overcoming these issues is crucial for advancing the field of irregular time series modeling.

\begin{figure}[!t]
  \centering  \includegraphics[width=\linewidth]{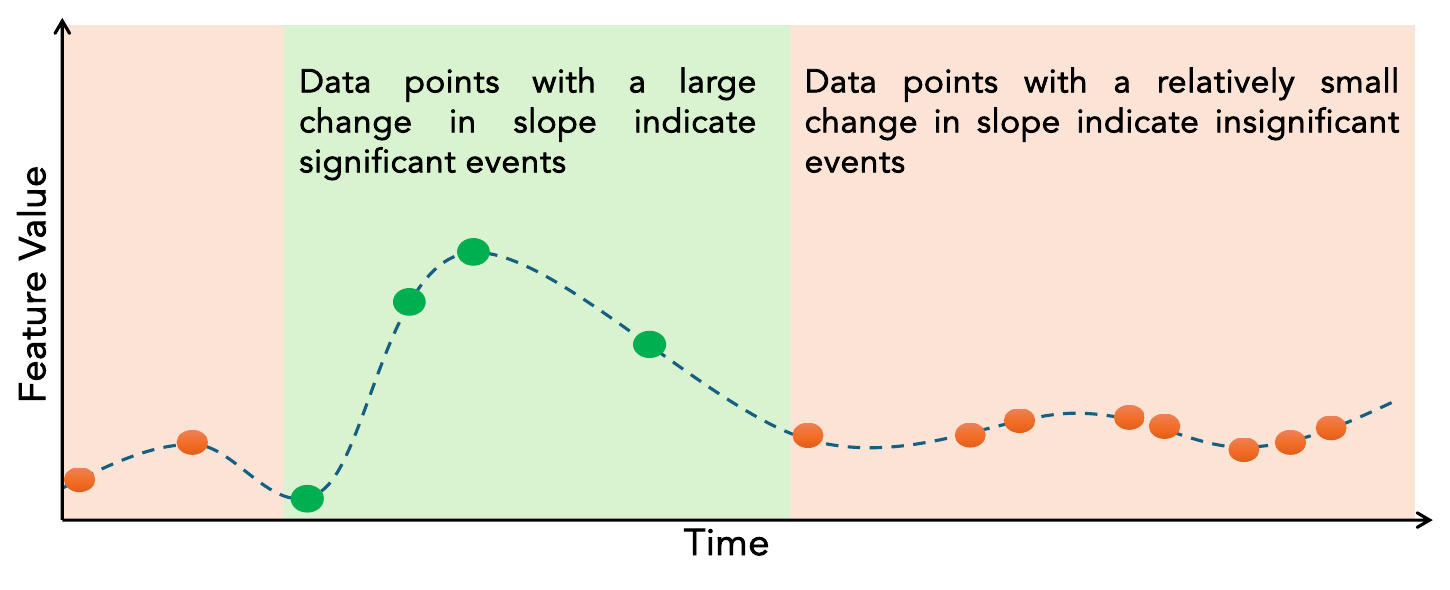}
\caption{An illustration of significant and insignificant events in the context of irregular time series. Points exhibiting a large rate of change are highlighted in green and are considered significant events. Conversely, points highlighted in orange have a relatively low rate of change and are considered insignificant events. Our model, EMIT, prioritizes masking and reconstruction of points associated with significant events, focusing on green regions that exhibit a large rate of change.}
  \label{fig:roc}
\end{figure}

To address these challenges, we can leverage a key characteristic of irregular time series: the rate of change as shown in Fig.~\ref{fig:roc}. Understanding how data points change over time provides valuable insights. Therefore, this paper proposes a new pretraining framework with a masked autoencoder specifically designed for irregular time series based on the rate of change. Our approach masks points with a higher rate of change, which typically indicate significant or noteworthy events in the time series, and then attempts to reconstruct them. Instead of masking raw inputs, we mask the embeddings of these elements (time, value, feature) and aim to reconstruct the embeddings. Masking the embeddings rather than raw data allows the model to develop features that are inherently robust and generalized. Our framework is versatile and can be integrated with any encoder that explicitly handles time, features, and values. Furthermore, we demonstrate through experiments on two benchmark datasets (Mimic-III~\cite{mimic_iii} and PhysioNet Challenge 2012~\cite{physionet_2012} that the high-quality features learned through our pretraining task can be effectively fine-tuned with a limited set of labelled data. Our main contributions are summarized below:

\begin{itemize}
    \item This paper introduces a framework for pretraining irregular time series data by focusing on masking-based reconstruction in the latent space.
    \item This paper proposes an algorithm to create masks based on the rate of change in variables over time.
    \item Extensive experimentation using two benchmark datasets shows that EMIT can improve the quality of representations of irregular time series.
\end{itemize}

The rest of the paper is organized as follows: Section~\ref{sec:related-work} reviews existing literature on irregular time series learning and explores self-supervised learning methodologies in both regular and irregular time series. In Section~\ref{sec:emit-frame}, we formulate the problem and describe the EMIT framework. Section~\ref{sec:exp} presents experimental results and relevant ablation studies. Section~\ref{sec:conclusion} concludes the paper and suggests directions for future research.

\section{RELATED WORK}
\label{sec:related-work}
\subsection{Irregular Time Series Learning}
Early studies demonstrated the potential of RNN-based models for handling such data. For instance, Lipton and others in 2015 showed how LSTMs could effectively model clinical time series data by imputing missing entries through heuristic methods~\cite{lipton2015learning}. Building on this, a method is introduced to enhance LSTM performance by using binary indicators to directly incorporate missing data as features~\cite{lipton2016directly}. Customizing architectures for irregular intervals led to further improvements. T-LSTM, for example, modified LSTM’s memory cells to diminish the influence of past data if the intervals between observations were long~\cite{T-LSTM}. Similarly, GRU-D introduced trainable decays for input and hidden states to maintain relevance to the current input based on the time elapsed~\cite{GRU-D}. Additionally, IP-NET proposed an end-to-end framework where an interpolation network first processes sparse data relative to reference points, and a prediction network then consumes its output to forecast future states~\cite{IP-NET}. Although effective, these methods, by discretizing latent dynamics, risk losing some information about continuous changes in the system. To mitigate this, ODE-RNN modified RNNs by incorporating differential equations to model the continuous-time dynamics of latent states, thus outperforming traditional RNNs in irregular time series contexts~\cite{Rubanova2019LatentOD}.
However, RNNs have scalability issues as time series length increases. Drawing inspiration from transformers~\cite{transformer}, the SeFT model shifted the perspective by treating time series analysis more like a set classification problem, which allowed for parallel processing and improved scalability for longer sequences~\cite{seft}.

\subsection{Self-supervised Learning}
\subsubsection{Self-supervised learning in general machine learning}
Supervised representation learning often requires manual labelling, which is time-consuming and expensive~\cite{imagenet,CheXpert}. To address these challenges, Self-Supervised Learning (SSL) became popular where a part of input was used to provide supervisory signals. SSL observed a big success in CV~\cite{moco, MAE, simclr, BYOL, SWaV} and NLP~\cite{GPT,BERT, roberta}, which inspired its adoption to other domains like time series. 

\subsubsection{Self-supervised learning in regular time series}
Regular time series, characterized by consistent intervals between measurements, benefit from various SSL techniques, as detailed in surveys by~\cite{timeseries_SSL_survey, PTM_TS_Review}. TimeNET~\cite{timenet} leverages a masked autoencoder to learn dense vector representations from time series of varying lengths. Triplet loss with time-based negative sampling is introduced by~\cite{triplet_loss}. TST~\cite{TST} employs a BERT-like~\cite{BERT} strategy using transformers to reconstruct masked values, replacing input values with zeros. TNC~\cite{TNC} applies a contrastive objective inspired by positive unlabelled learning, differentiating neighboring from non-neighboring samples. TS-TCC~\cite{TS-TCC} utilizes dual contrastive mechanisms—temporal and contextual—to develop robust and discriminative representations. TARNet~\cite{tarnet} alternates between pretext and downstream tasks, using task-informed data masking. Multimodal clinical time series contrastive learning with structured and physiological data is explored in~\cite{multimodal_clincial_time_series_cl}. BTSF~\cite{btsf} employs unique augmentation and bilinear fusion techniques to enhance feature representations by integrating time and spectral data. TF-C~\cite{TF-C} focuses on integrating time-frequency components for SSL. SimMTM~\cite{simmtm} proposes a framework inspired by manifold learning to reconstruct the original series through the weighted aggregation of neighboring masked series. FOCAL~\cite{focal} extracts shared and distinct features from different modalities for robust SSL. InfoTS \cite{infots} uses a meta learner to select augmentations based on fidelity and variability. MHCCL \cite{MHCCL} enhances contrastive learning with bidirectional masking in hierarchical clustering partitions. BasisFormer~\cite{basisformer} adapts contrastive learning to extract basis from historical and future sections of a time series. Lastly, COMET~\cite{contrast_everything} implements hierarchical contrasting at various levels within medical time series, from patient to observation.

Although these methods have shown promising results in regular time series, the unique challenges posed by irregular time series, such as varying sampling intervals require the development of specialised SSL techniques tailored to handle these complexities.

\subsubsection{Self-supervised learning in irregular time series}
Researchers have explored several self-supervised strategies for irregular time series. These include window-based forecasting, contrastive learning, and masking-based reconstruction. STraTS~\cite{strats} used forecasting to pretrain their model, leading to improved performance compared to models without a pretext task. PrimeNET~\cite{primenet} employed two time-sensitive pretext tasks: time-aware augmentations for contrastive learning and constant time masking. In this paper, we leverage a key characteristic of irregular time series, the rate of change, to create a more effective masking strategy.

\begin{figure*}[!t]
  \centering
  \includegraphics[width=0.8\textwidth]{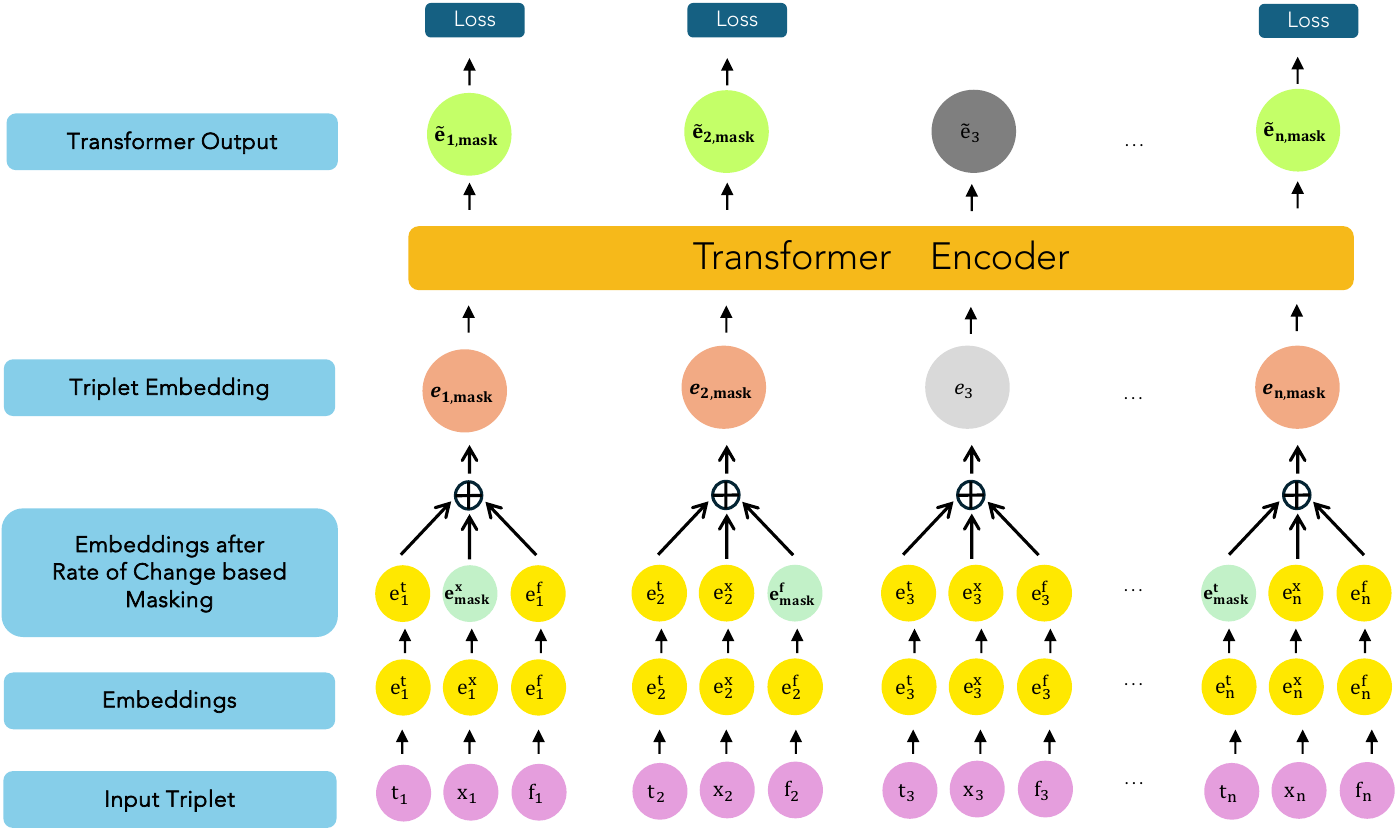}
  \caption{EMIT pretraining architecture. The initial input triplets are embedded and subsequently masked with the respective masking token. The masks are selected based on events identified by their rate of change as described in Algorithm~\ref{alg:ev-mask} and~\ref{alg:rate}. The embeddings are then summed to produce the final triplet embedding, which is fed into the transformer encoder blocks. The transformer attempts to reconstruct the masked embeddings using the remaining unmasked embeddings. The generated embeddings are then used for loss calculation.}
  \label{fig:architecture}
\end{figure*}

\section{EMIT Framework}

\label{sec:emit-frame}
\subsection{Problem Formulation}
Our dataset consists of irregular time series represented by a sequence of observation triplets:
\begin{equation}
T = \{(t_i, x_i, f_i) \mid i = 1, \ldots, n\},
\end{equation}
where \( t_i \in \mathbb{R}_{\geq 0} \) (non-negative real numbers), \( x_i \in \mathbb{R} \) (real numbers), and \( f_i \in F \) (a set of categorical features). Each \( T \) is a time series comprising these triplets. The dataset \( D \) consists of \( N \) such time series, each paired with a corresponding binary label:
\begin{equation}
    D = \{(T_j, y_j) \mid y_j \in \{0, 1\}, j = 1, \ldots, N \}.
\end{equation}

\textbf{Problem:} The primary objective is to learn a model \( P \) from the given unlabelled pretraining dataset \( D_{pt} \) containing \( N \) sequences. The model should generate generalizable representations for specific downstream tasks. 

Our framework, EMIT, can be implemented with any architecture that explicitly encodes temporal, feature, and value information. In this work, we employ an architecture inspired by STraTS~\cite{strats} as our backbone model.

\subsection{Model Architecture}
Our model architecture comprises input embeddings, a transformer encoder, a time series embedding mechanism, a projection head, and a self-supervised pretext task for learning representations from irregular time series data. Fig. \ref{fig:architecture} illustrates major components of our pretraining architetcure.

\subsubsection{Input Embeddings}
For the input time series \( T \), we obtain input embeddings by combining embeddings for time, value, and features for each observation:
\begin{equation}
\mathbf{e_i} = \mathbf{e}_i^t + \mathbf{e}_i^x + \mathbf{e}_i^f. 
\label{eq:sum_mask}
\end{equation}

Here, \(\mathbf{e_i} \in \mathbb{R}^d\), \(\mathbf{e}_i^t \in \mathbb{R}^d\), \(\mathbf{e}_i^x \in \mathbb{R}^d\), and \(\mathbf{e}_i^f \in \mathbb{R}^d\) denote the embeddings for the combined input, time, value, and feature, respectively. The time and value embeddings are derived from a Feed-Forward Network(FFN). FFN is a single hidden layer neural network with $tanh$ activation:
\begin{align}
\mathbf{e}_i^t &= \text{FFN}_{\text{time}}(t_i),\\
\mathbf{e}_i^x &= \text{FFN}_{\text{value}}(x_i).
\end{align}
The feature embeddings are obtained using an embedding matrix:
\begin{equation}
\mathbf{e}_i^f = \text{Embed}(f_i).
\end{equation}

\subsubsection{Transformer Encoder}

Following the methods described in~\cite{transformer,strats,contiformer, peng2021sequential}, each embedded observation \( \mathbf{e}_i \) from the set 
\(\{\mathbf{e}_1, \ldots, \mathbf{e}_n\}\) is processed through a series of \( m \) transformer encoder blocks, each containing \( h_e \) attention heads. The output from one block feeds sequentially into the next, resulting in a series of refined embeddings:
\begin{equation}
\{\tilde{\mathbf{e}}_1, \ldots, \tilde{\mathbf{e}}_n\} = \text{TrEnc}(\{\mathbf{e}_1, \ldots, \mathbf{e}_n\}),
\end{equation}
where each \(\tilde{\mathbf{e}}_i \in \mathbb{R}^d\).

These contextualized embeddings integrate temporal and feature-based information, augmented by the multi-head attention mechanism, thereby offering a refined understanding of the input sequence.

\subsubsection{Time Series Embedding}

To synthesize a comprehensive embedding for the entire time series, we use an attention-based aggregation mechanism:
\begin{align}
a_i &= \mathbf{u}_a^\top \tanh(\mathbf{W}_a \tilde{\mathbf{e}}_i + \mathbf{b}_a), \\
\alpha_i &= \text{softmax}(a_i) \quad \text{for} \quad i = 1, \ldots, n,
\end{align}
where \( \mathbf{W}_a \in \mathbb{R}^{d_a \times d} \), \( \mathbf{b}_a \in \mathbb{R}^{d_a} \), and \( \mathbf{u}_a \in \mathbb{R}^{d_a} \).

The attention weights \( \alpha_i \) determine the significance of each context-aware embedding \( \tilde{\mathbf{e}}_i \). These weights are then used to construct an aggregated time series embedding:
\begin{equation}
\mathbf{e}_T = \sum_{i=1}^{n} \alpha_i \tilde{\mathbf{e}}_i.
\end{equation}

This composite embedding \( \mathbf{e}_T  \in \mathbb{R}^d\) is a time series representation that can then be used for downstream tasks.

\subsubsection{Prediction Head}

To facilitate predictions for downstream tasks, the aggregated time series embedding is passed through a simple FFN, which comprises a dense layer followed by a sigmoid activation function:
\begin{equation}
\tilde{y} = \text{sigmoid}(\mathbf{w}_o^\top \mathbf{e}_T + \mathbf{b}_o),
\end{equation}
where \( \mathbf{w}_o \in \mathbb{R}^d \) represents the weights of the dense layer, and \( \mathbf{b}_o \in \mathbb{R} \) is the bias term.

\subsection{Model Pretraining}
In this section, we describe our model pretraining strategy, which is based on masking-based reconstruction, where the masks are selected according to the rate of change.
\subsubsection{Selection of Masks}

We begin with the calculation of the rate of change for data points across the time series, which is crucial for identifying significant fluctuations in data values over time. The rate of change \( r_i \) is defined as:
\begin{equation}
r_i = \frac{X_{i+1} - X_i}{T_{i+1} - T_i},
\end{equation}
where \( X_i \) represents the value at the \( i \)-th time point, and \( T_i \) denotes the corresponding time. This equation helps in determining how quickly data values change between consecutive points. The algorithm for calculating the rate of change is shown in Algorithm~\ref{alg:rate}.

To determine whether a change is significant, we compare the computed rate of change to a predefined threshold \( \theta \):
\begin{equation}
\text{Significance Criterion}: |r_i| > \theta.
\end{equation}

This condition identifies critical changes within the data, suggesting potentially influential fluctuations that the model needs to focus on.

After identifying the significant and less significant events, we apply different probabilities for masking these events to focus the model's learning effectively, as described in Algorithm~\ref{alg:ev-mask}:
\begin{equation}
M_i = \begin{cases} 
1 & \text{if } |r_i| > \theta \text{ and } p < (1 - \alpha) \\
1 & \text{if } |r_i| \leq \theta \text{ and } p < \alpha \\
0 & \text{otherwise}
\end{cases},
\end{equation}
where \(M_i\) represents boolean mask at the i-th time point. The \( p \) is a random variable uniformly distributed between 0 and 1. The parameter \( \alpha \) denotes the probability threshold for masking less significant changes, and \( (1 - \alpha) \) is used for masking significant changes.

Applying masks to both significant and less significant changes ensures comprehensive model training, providing balanced exposure to both significant changes and regular patterns. This enhances the model's robustness, preparing it for a variety of real-world scenarios.

\subsubsection{Applying Masks to Embeddings}
We use specific mask tokens for time, value, and feature, denoted as \( \mathbf{e}_{\text{mask}}^t \), \( \mathbf{e}_{\text{mask}}^x \), and \( \mathbf{e}_{\text{mask}}^f \), respectively. The shapes of these mask tokens are \( \mathbb{R}^d \).

For each observation, we randomly choose to mask any of the time, value, or feature embedding. The final embedding \( \mathbf{e}_{\text{i, mask}} \) is obtained by combining the masked embedding with the unmasked ones:
\begin{align}
\mathbf{e}_{\text{i, mask}} &= \mathbf{e}_{\text{mask}}^t + \mathbf{e}_i^x + \mathbf{e}_i^f \quad \text{if time is masked} \label{eq:time_mask}, \\
\mathbf{e}_{\text{i, mask}} &= \mathbf{e}_i^t + \mathbf{e}_{\text{mask}}^x + \mathbf{e}_i^f \quad \text{if value is masked} \label{eq:value_mask}, \\
\mathbf{e}_{\text{i, mask}} &= \mathbf{e}_i^t + \mathbf{e}_i^x + \mathbf{e}_{\text{mask}}^f \quad \text{if feature is masked} \label{eq:feature_mask},
\end{align}
\begin{equation}
\begin{aligned}
\{\tilde{\mathbf{e}}_1, \ldots, \tilde{\mathbf{e}}_{\text{i, mask}}, \ldots, \tilde{\mathbf{e}}_n\} & = \\
\operatorname{TrEnc}(\{\mathbf{e}_1, \ldots, \mathbf{e}_{\text{i, mask}}, \ldots, \mathbf{e}_n\})
\end{aligned}
\end{equation}

\subsubsection{Objective Function}

\textbf{Reconstruction Loss:} For each masked position \( i \) (where \( M_i = 1 \)), the output embedding \( \mathbf{\tilde{e}}_{\text{i, mask}} \) is compared to the original embedding \( \mathbf{e}_{\text{i, mask}} \) using Mean Squared Error (MSE):
\begin{equation}
L_{\text{recon}} = \frac{1}{|M|} \sum_{i=1}^{n} M_i \cdot \frac{1}{d} \sum_{k=1}^{d} (\mathbf{\tilde{e}}_{\text{i, mask}}^k - \mathbf{e}_{\text{i, mask}}^k)^2.
\end{equation}

\textbf{Forecasting Loss:} We also employ a forecasting loss as an auxiliary objective. For the forecasting component, we use the same masked embeddings \( \mathbf{\tilde{e}}_{\text{i, mask}} \). The forecasting layer is a FFN with \( F \) neurons:
\begin{align}
\hat{y}_i &= \text{ForecastingLayer}(\mathbf{\tilde{e}}_{\text{i, mask}}), \\       
L_{\text{forecast}} &= \text{forecast\_loss}(y_i, \hat{y}_i).
\end{align}

\textbf{Total Loss:} The total loss combines the reconstruction loss and the forecasting loss, weighted by an error coefficient \( \lambda \):
\begin{equation}
L_{\text{total}} = L_{\text{forecast}} + \lambda \cdot L_{\text{recon}}.
\end{equation}

\begin{algorithm}[!t]
\caption{Event-based Masking for a Sequence}
\label{alg:ev-mask}
\begin{algorithmic}[1]
\Require $T$: Array of times, $X$: Array of values, $F$: Array of features, $\theta$: Threshold for significant rate of change, $\alpha$: Probability for less significant changes, $1 - \alpha$: Probability for significant changes
\Ensure $M$: Mask array indicating points to be masked
\State Initialize $M$ as a boolean array of zeros with length equal to $X$
\For{each unique feature $f$ in $F$}
    \State $I \leftarrow \{i \mid F_i = f\}$ \Comment{Indices of entries in feature $f$}
    \If{$|I| > 1$}
        \State $R \leftarrow \text{calculate\_rate\_of\_change}(X[I], T[I])$
        \State $S \leftarrow \{j \mid |R_j| > \theta\}$ \Comment{Significant changes}
        \State $U \leftarrow \{j \mid |R_j| \leq \theta\}$ \Comment{Insignificant changes}
        \For{$j \in S$}
            \State $M_{I[j]} \leftarrow \text{uniform}(0, 1) < (1 - \alpha)$ \Comment{Masking probability for significant changes}
        \EndFor
        \For{$j \in U$}
            \State $M_{I[j]}  \leftarrow \text{uniform}(0, 1) < \alpha$ \Comment{Masking probability for insignificant changes}
        \EndFor
    \EndIf
\EndFor
\State \Return $M$
\end{algorithmic}
\end{algorithm}

\begin{algorithm}[!t]
\caption{Calculate Rate of Change}
\label{alg:rate}
\begin{algorithmic}[1]
\Require $X$: Array of values, $T$: Array of corresponding times
\Ensure $R$: Array of rates of change
\State $\Delta T \leftarrow \text{diff}(T)$ \Comment{Calculate differences between consecutive time points}
\State $\Delta X \leftarrow \text{diff}(X)$ \Comment{Calculate differences between consecutive values}
\State $R \leftarrow \text{zeros}(\text{length}(X))$ \Comment{Initialize rates with zeros}
\For{$i \leftarrow 1$ to \text{length}($\Delta T$)}
    \If{$\Delta T_i \neq 0$}
        \State $R_i \leftarrow \Delta X_i / \Delta T_i$
    \Else
        \State $R_i \leftarrow 0$ \Comment{Handle division by zero}
    \EndIf
\EndFor
\State $R_\text{length(R)} \leftarrow 0$ \Comment{Set the last element to zero}
\State \Return $R$
\end{algorithmic}
\end{algorithm}

\section{Experiment}
\label{sec:exp}
In this section, we describe the experimental settings and results, aiming to address the following research questions (RQs):
\begin{itemize}
\item RQ1: How does EMIT perform in comparison with state-of-the-art irregular time series representation learning models?
\item RQ2: How effective is the event mask compared to the random mask?
\item RQ3: What is the impact of masking different embeddings on the results?
\item RQ4: How effective are various self-supervised learning strategies for irregular time series?
\item RQ5: How do hyperparameter settings influence the performance of EMIT?
\end{itemize}

\subsection{Setup}

\subsubsection{Datasets}
We assessed our proposed framework using two clinical datasets. Initially, we pretrained our model, followed by fine-tuning with various fractions of labelled data for the mortality prediction task. We adopted a similar methodology to the STraTS\cite{strats} for preprocessing and preparing the datasets to ensure a fair comparison. The \textbf{MIMIC-III} dataset, which is publicly available, includes electronic health records from over 40,000 patients in critical care units at Beth Israel Deaconess Medical Center, spanning from 2001 to 2012. The \textbf{ PhysioNet Challenge 2012}\footnote{https://physionet.org/content/challenge-2012/1.0.0/}~\cite{physionet_2012} dataset contains EHRs from more than 10,000 ICU stays and was used for a mortality prediction challenge in 2012. We excluded demographic information from our analysis to avoid potential biases and inaccuracies that could arise from imputation, given that these details might not be available for all patients. Table~\ref{table:dataset_comparison} highlights key statistics for both the datasets.

\begin{table}[!t]
\centering
\caption{Dataset statistics. Adapted from \cite{strats}}
\resizebox{\linewidth}{!}{
\begin{tabular}{l|c|c}
    \toprule
    \textbf{Attribute} & \textbf{MIMIC-III} & \textbf{PhysioNet-2012} \\
    \midrule
    \# ICU stays & 52,871 & 11,988 \\
    \# ICU stays (with labels) & 44,812 & 11,988 \\
    Avg. span of time-series (with labels) & 23.5h & 47.3h \\
    \# Clinical variables & 129 & 37 \\
    Avg. variable missing rate & 89.7\% & 79.7\% \\
    Downstream Task & 24-hour mortality & 48-hour mortality \\
    \% positive class & 9.7\% & 14.2\% \\
    \bottomrule
\end{tabular}
}
\label{table:dataset_comparison}
\end{table}

\subsubsection{Metrics}
We used several evaluation metrics to assess our model's performance on the mortality prediction task compared to baseline models. These metrics include ROC-AUC (Area Under the Receiver Operating Characteristic Curve), PR-AUC (Area Under the Precision-Recall Curve), and min(Re, Pr), which represents the highest of the lowest values between recall and precision. ROC-AUC is commonly used in general classification tasks, whereas PR-AUC and min(Re, Pr) are especially beneficial for addressing challenges associated with imbalanced datasets.

\begin{table*}[!t]
\centering
\caption{Overall performance on MIMIC-III and PhysioNet-2012 on 50\% labelled data. Bold numbers indicate the best performance of all methods. Some baselines results are adopted from \cite{strats} due to similar experimental setup.}
\label{tab:overall}
\resizebox{0.97\linewidth}{!}{

\begin{tabular}{l|ccc|ccc}
\toprule
\multirow{2}{*}[-0.5ex]{Methods} & \multicolumn{3}{c|}{MIMIC-III} & \multicolumn{3}{c}{PhysioNet-2012} \\
\cmidrule(lr){2-4} \cmidrule(lr){5-7}
 & ROC-AUC & PR-AUC & min(Re, Pr) & ROC-AUC & PR-AUC & min(Re, Pr) \\
\midrule

\multicolumn{7}{l}{RNN-based} \\
GRU & 0.886 ± 0.002 & 0.559 ± 0.005 & 0.533 ± 0.007 & 0.831 ± 0.003 & 0.468 ± 0.008 & 0.465 ± 0.009 \\
GRU-D & 0.883 ± 0.003 & 0.544 ± 0.007 & 0.527 ± 0.005 & 0.833 ± 0.005 & 0.481 ± 0.008 & 0.468 ± 0.012 \\
InterpNet & 0.881 ± 0.002 & 0.540 ± 0.007 & 0.516 ± 0.005 & 0.822 ± 0.007 & 0.460 ± 0.017 & 0.455 ± 0.017 \\
\midrule
\multicolumn{7}{l}{CNN-based} \\
TCN & 0.879 ± 0.001 & 0.540 ± 0.004 & 0.525 ± 0.005 & 0.813 ± 0.005 & 0.430 ± 0.010 & 0.433 ± 0.009 \\
\midrule
\multicolumn{7}{l}{Transformer-based} \\
SaND & 0.876 ± 0.002 & 0.533 ± 0.011 & 0.515 ± 0.008 & 0.800 ± 0.013 & 0.406 ± 0.021 & 0.418 ± 0.018 \\
SeFT & 0.881 ± 0.003 & 0.547 ± 0.011 & 0.524 ± 0.010 & 0.832 ± 0.005 & 0.454 ± 0.017 & 0.465 ± 0.009 \\
STraTS & 0.885 ± 0.003 & 0.558 ± 0.011 & 0.532 ± 0.008 & 0.828 ± 0.006 & 0.478 ± 0.012 & 0.466 ± 0.014 \\
\midrule
\multicolumn{7}{l}{Ours} \\
\rowcolor{lightgray}
EMIT  & \textbf{0.891 ± 0.001} & \textbf{0.578 ± 0.001} & \textbf{0.539 ± 0.004} & \textbf{0.846 ± 0.002} & \textbf{0.506 ± 0.015} & \textbf{0.486 ± 0.010} \\
\bottomrule
\end{tabular}}
\end{table*}
\hfill
\subsubsection{Baselines} \hfill \break
\vspace{-1em} % Optional: adjust vertical space as needed

\textbf{RNN-based Models}
\begin{itemize}
    \item \textbf{Gated Recurrent Unit (GRU)}\cite{GRU}: To address irregularity, this model uses hourly aggregation of time-series data, employing mean imputation for missing values. It includes additional features such as indicators of data missingness and time intervals between observations.
    \item \textbf{GRU with Decay Mechanism (GRU-D)}\cite{GRU-D}: This is a variant of the GRU model that incorporates trainable decay factors to handle the decay of information over time.
    \item \textbf{Interpolation-prediction Network (InterpNet)}\cite{IP-NET}: A semi-parametric interpolation network designed to interpolate features to fixed reference time points. These interpolated data points are then used as input to a GRU-based prediction network.
\end{itemize}

\textbf{Convolution-based Models}
\begin{itemize}
    \item \textbf{Temporal Convolutional Network (TCN)}\cite{TCN}: This network utilizes dilated convolutions to expand the receptive field, along with residual connections to aid in learning across deeper structures.
\end{itemize}

\textbf{Transformer-based Models}
\begin{itemize}
    \item \textbf{Simply Attend and Diagnose (SaND)}\cite{sand}: Features a masked self-attention mechanism that focuses on segments of the input sequence, using positional encoding to maintain the temporal order and dense interpolation strategies to handle irregular sampling intervals.
    \item \textbf{Set Functions for Time Series (SeFT)}\cite{seft}: Represents the input data as sets of observation triplets (time, feature, value). This model uses multi-head attention to process these sets and employs aggregation techniques to synthesize the outputs.
    \item \textbf{Self-supervised Transformer for Time-Series (STraTS)}\cite{strats}: A variant of SeFT that introduces a novel embedding strategy. It employs a transformer-like attention mechanism and aggregation to enhance the handling of time-series data.
\end{itemize}

\hfill
\vspace{-0.25em}
\subsubsection{Implementation}
For both datasets, a Transformer Encoder~\cite{transformer} was used for encoding. For fair evaluation, the hyperparameters of the Transformer model were kept similar to STraTS\cite{strats}: 2 encoder blocks and 4 attention heads. Our scientific hyperparameters - rate of change threshold was chosen from \{1e-1, 1e-2, 1e-3\}, and the probability for masking insignificant events was chosen from \{0, 0.1, \ldots, 1.0\}.

% In EMIT pretraining, for the MIMIC dataset, the batch size was \{256\}, while for PhysioNet it was \{128\}, learning rate was \{5e-4\}, and dropout rates was \{0.2\}~\cite{dropout}. Using Adam~\cite{adam_opt} as the default optimizer, the learning rate weight decay value was set to \{0\}. For MIMIC, the time series length was 880 and the embedding dimension was 50. For PhysioNet, the maximum time series length was 791 and the embedding dimension was also 50. The maximum length was chosen as advised by STraTS. The models were pretrained for 100 epochs with early-stop. All experiments were conducted on a single GPU, from H100, A100, V100, and L40, with the training time per run varied from 4 to 7 hours.

In EMIT pretraining, the MIMIC dataset used a batch size of {256} and the PhysioNet dataset used {128}. Both datasets had a learning rate of {5e-4} and a dropout rate of {0.2}\cite{dropout}. The Adam optimizer\cite{adam_opt} was used with a weight decay of {0}. For MIMIC, the time series length was 880 with an embedding dimension of 50; for PhysioNet, the maximum time series length was 791, also with an embedding dimension of 50, as advised by STraTS\cite{strats}. Models were pretrained for 100 epochs with early stopping. All experiments were conducted on a single GPU from H100, A100, V100, or L40, with training times ranging from 4 to 7 hours.

% For fine-tuning, for both datasets, the batch size was \{32\}, the learning rate was \{5e-5\}, the dropout rates were \{0.2, 0.4\}, and the weight decay values were \{0, 1e-4\}. The models were fine-tuned for 100 epochs with early-stop. The fine-tuning time per model varies from 1 to 3 hours. These experiments were repeated 5 times.

For fine-tuning, both datasets used a batch size of 32, a learning rate of 5e-5, dropout rates of \{0.2, 0.4\}, and weight decay values of \{0, 1e-4\}. Models were fine-tuned for 100 epochs with early stopping, and the fine-tuning time per model ranged from 1 to 3 hours. These experiments were repeated 5 times.

\begin{figure*}[!t]
  \centering
  \includegraphics[width=0.99\linewidth]{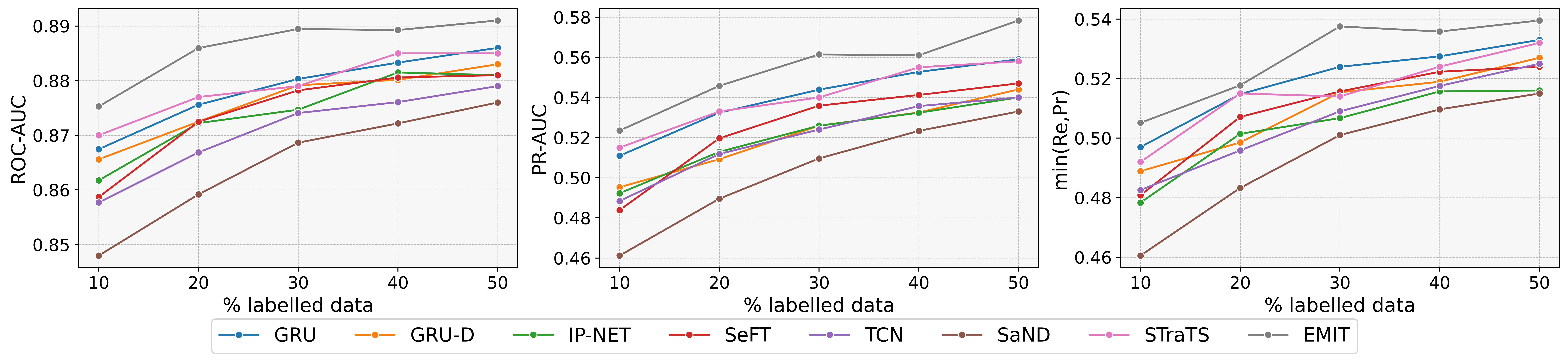}
  \caption{Mortality prediction performance of different baseline models on mimic dataset trained at different percentage of labelled data.} 
 \label{fig:overall_results_different_models_mimic}
\end{figure*}

\begin{figure*}[!t]
  \centering
  \includegraphics[width=0.99\linewidth]{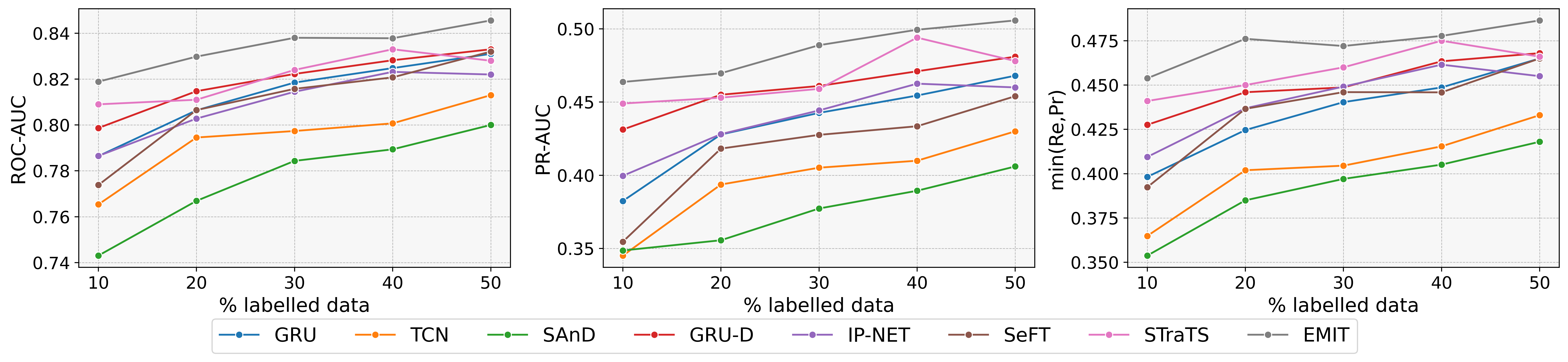}
  \caption{Mortality prediction performance of different baseline models on PhysioNet Challenge 2012 dataset trained at different percentage of labelled data.} 
 \label{fig:overall_results_different_models_PhysioNet}
\end{figure*}

\subsection{Overall Performance (RQ1)}

To evaluate the overall performance of our proposed method, EMIT, we conducted experiments on two healthcare datasets: MIMIC-III and PhysioNet-2012. We compared EMIT against several baseline models, including RNN-based models \cite{GRU, GRU-D, IP-NET}, convolution-based models\cite{TCN} and transformer-based models \cite{sand, seft, strats}. The evaluation metrics used were ROC-AUC, PR-AUC, and the minimum of recall and precision (min(Re, Pr)) across all possible decision thresholds.

\begin{table*}[!t]
\centering
\caption{Ablation studies on MIMIC-III and PhysioNet-2012. Bold Numbers indicate the highest value for a given metric. }\label{tab:ablations}
\resizebox{0.99\linewidth}{!}{ %
\begin{tabular}{l|ccc|ccc}
\toprule
\multirow{2}{*}[-0.5ex]{Methods} & \multicolumn{3}{c}{MIMIC-III} & \multicolumn{3}{c}{PhysioNet-2012} \\
\cmidrule(lr){2-4} \cmidrule(lr){5-7}
& ROC-AUC & PR-AUC & min(Re, Pr) & ROC-AUC & PR-AUC & min(Re, Pr) \\
\midrule

Random mask & 0.885 ± 0.001 & 0.558 ± 0.005 & 0.532 ± 0.003 & 0.841 ± 0.003 & 0.494 ± 0.002 & 0.476 ± 0.004 \\
\midrule
Time EV mask & 0.888 ± 0.002 & 0.571 ± 0.004 & \textbf{0.542 ± 0.006} & 0.837 ± 0.004 & 0.485 ± 0.010 & 0.478 ± 0.009 \\
Value EV mask & 0.886 ± 0.002 & 0.561 ± 0.010 & 0.531 ± 0.000 & 0.840 ± 0.002 & 0.478 ± 0.006 & 0.481 ± 0.004 \\
Feature EV mask & 0.889 ± 0.003 & 0.569 ± 0.003 & 0.531 ± 0.001 & 0.838 ± 0.002 & 0.489 ± 0.000 & 0.461 ± 0.006 \\
Emb EV mask & 0.884 ± 0.002 & 0.556 ± 0.001 & 0.538 ± 0.006 & 0.834 ± 0.001 & 0.470 ± 0.012 & 0.459 ± 0.011 \\
\midrule
\rowcolor{lightgray}
EMIT  & \textbf{0.891 ± 0.001} & \textbf{0.578 ± 0.001} & 0.539 ± 0.004 & \textbf{0.846 ± 0.002} & \textbf{0.506 ± 0.015} & \textbf{0.486 ± 0.010} \\

\bottomrule
\end{tabular}}
\end{table*}

\begin{table*}[!t]\centering
\caption{Different self-supervised learning strategies.Bold numbers indicate the best performance of all methods.}\label{tab:different_SSL}
\scriptsize %
\resizebox{0.99\linewidth}{!}{ %
\begin{tabular}{l|ccc|ccc}
\toprule
\multirow{2}{*}[-0.5ex]{SSL} &\multicolumn{3}{c}{MIMIC-III} &\multicolumn{3}{c}{PhysioNet-2012}\\
\cmidrule(lr){2-4} \cmidrule(lr){5-7}
&ROC-AUC &PR-AUC &min(Re, Pr) &ROC-AUC &PR-AUC &min(Re, Pr) \\\midrule

Forecasting &0.885 ± 0.003 & 0.558 ± 0.011 & 0.532 ± 0.008 & 0.828 ± 0.006 & 0.478 ± 0.012 & 0.466 ± 0.014 \\
Time CL  & 0.871 ± 0.001 & 0.527 ± 0.003 & 0.504 ± 0.005 & 0.811 ± 0.011 & 0.408 ± 0.021 & 0.424 ± 0.021\\
\midrule
\rowcolor{lightgray}
EMIT  & \textbf{0.891 ± 0.001} & \textbf{0.578 ± 0.001} & \textbf{0.539 ± 0.004} & \textbf{0.846 ± 0.002} & \textbf{0.506 ± 0.015} & \textbf{0.486 ± 0.010} \\

\bottomrule
\end{tabular}}
\end{table*}

The results, summarized in Table~\ref{tab:overall}, highlight several key insights. EMIT consistently achieved higher scores on both datasets across each metric, demonstrating its superior ability to balance precision and recall when identifying mortality risk events, which are the minority class. RNN-based models like GRU and GRU-D, although performing reasonably well, struggled with capturing long-range dependencies and complex temporal patterns due to their sequential nature. Convolution-based models, such as TCN, excelled at capturing local temporal structures but had difficulty modeling irregular sampling intervals and missing values typical in physiological data. Transformer-based models like SaND and SeFT, despite their strengths in other domains, lacked the necessary inductive biases to manage the unique characteristics of time series data effectively. While STraTS performed reasonably better than other baselines, it couldn't outperform EMIT. EMIT's success can be attributed to its novel event-based masking and reconstruction strategy, which allows it to focus on critical points of change within the time series data, thereby capturing significant patterns more accurately and improving predictive performance.
Additionally, to highlight the effectiveness, we finetuned the model on mortality prediction task with different percentages of labelled data ranging from 10\% to 50\%. As per Fig.~\ref{fig:overall_results_different_models_mimic} and Fig.~\ref{fig:overall_results_different_models_PhysioNet}, we can observe that EMIT can effectively learn from sparsely labelled scenarios and perform consistently better as labelled data increases, attributing its success to new self-supervision task.

\subsection{Ablation Studies (RQ2 \& RQ3)}

In the ablation study, the effectiveness of different masking strategies is comprehensively evaluated. First, we evaluated the importance of the event mask by replacing it with a random mask in EMIT. Second, within event masks, we evaluated the following variants: the \textbf{Time Event Mask} (Time EV Mask), described by Equation~\eqref{eq:time_mask}, which involves masking only the time embeddings for points identified by the event mask algorithm; the \textbf{Value Event Mask} (Value EV Mask), following Equation~\eqref{eq:value_mask}, which masks the value embeddings at critical changes identified by the event mask algorithm; the \textbf{Feature Event Mask} (Feature EV Mask), as per Equation~\eqref{eq:feature_mask}, which masks feature embeddings based on specific event classes identified by the algorithm; and the \textbf{Embedding Event Mask} (Emb EV Mask), applying Equation~\eqref{eq:sum_mask}, where the sum of the time, value, and feature embeddings is masked, targeting those identified by the event mask algorithm. The above methods were compared against, \textbf{EMIT}, a composite approach that dynamically selects between time, value, or feature embeddings for masking, based on the event positions identified by the algorithm.

From Table~\ref{tab:ablations}, it is evident that replacing random masks with event masks enhances model performance across both datasets. All event-based mask model variants outperform the random mask in all three metrics. This suggests that rate-of-change-based events are critical for learning representations in irregular time series.

Within event masking, it is observed that the model benefits more from masking time, value, or feature embeddings randomly, rather than masking them individually. Specifically, EMIT, which randomly chooses to mask either the time, value, or feature embedding, shows the best overall performance across the metrics and datasets. While time-only masking performs slightly better than EMIT in the min(Re, Pr) metric, it does not surpass EMIT in the other metrics or in the PhysioNet dataset. This indicates that the random selection of embeddings to mask, informed by event positions, provides a more comprehensive and effective approach for irregular time series representation learning.

\begin{figure*}[!t]
  \centering
  \includegraphics[width=0.99\textwidth]{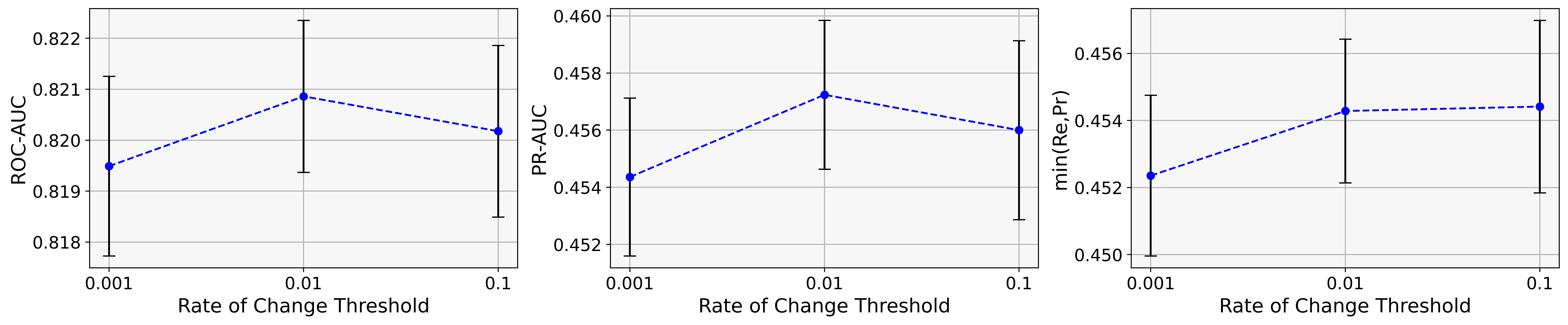}
  \caption{Effect of different rate of change threshold on prediction performance using PhysioNet Challenge 2012 dataset.} 
 \label{fig:sensitivity_analysis_rate_of_change_threshold_PhysioNet}
\end{figure*}

\begin{figure*}[!t]
  \centering
  \includegraphics[width=0.99\linewidth]{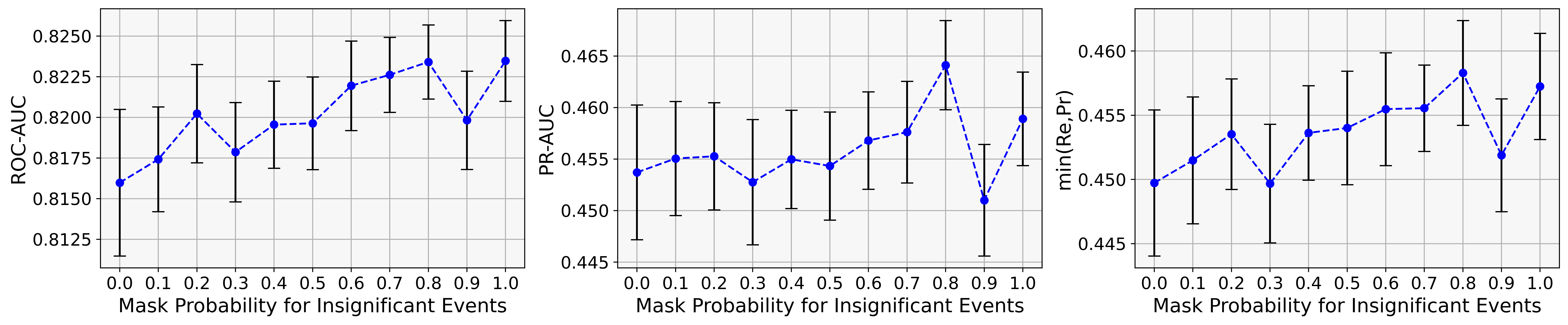}
  \caption{Effect of different mask probabilities of insignificant events on prediction performance using PhysioNet Challenge 2012 dataset.} 
 \label{fig:sensitivity_analysis_insignificant_event_probability_PhysioNet}
\end{figure*}

\subsection{Effectiveness of Different SSL on Irregular Time Series (RQ4)}

In this section, we compare the effectiveness of different pretraining methods applied to irregular time series data. The strategies examined include forecasting, time-sensitive contrastive learning, and EMIT. In the forecasting method, the model is trained on 24 hours of data to make predictions for the next 2-hour window, as used in STraTS~\cite{strats}. The time-sensitive contrastive learning approach generates positive and anchor pairs for a given time series, maintaining data density in both sparse and dense regions, as introduced in PrimeNET~\cite{primenet}. Table~\ref{tab:different_SSL} presents the performance of each method across the MIMIC-III and PhysioNet-2012 datasets. Our analysis reveals following key insights:

\textbf{Forecasting}: This method demonstrates strong performance, particularly in its ability to predict future values in irregular time series. However, it falls slightly behind in capturing the nuances of more complex temporal patterns compared to EMIT.
\textbf{Time-Sensitive Contrastive Learning} (Time CL): This approach shows promise but generally underperforms relative to the other methods. It struggles particularly with maintaining precision and recall in the PhysioNet-2012 dataset, indicating potential limitations in handling high variability in time series data.
\textbf{EMIT}: This method consistently outperforms the others, demonstrating superior ability to handle the complexities of irregular time series. By randomly choosing to mask time, value, or feature embeddings based on event detection, EMIT captures critical temporal dynamics more effectively. Its strong performance across all evaluated metrics suggests that it provides a more robust framework for pretraining models on irregular time series data.
These findings highlight the importance of selecting appropriate pretraining strategies for irregular time series. While traditional forecasting methods are reliable, advanced techniques like EMIT offer significant improvements by effectively leveraging the underlying event-based structure of the data.

\subsection{Hyperparameter sensitivity (RQ5)}
In this section, we describe the effect of our two scientific hyperparameters on the performance of mortality prediction task from PhysioNet 2012 dataset. These two parameter were rate of change threshold, which controls if the given rate of change is significant or not and second hyperparameter is probability at which insignificant event will be masked.

From Fig.~\ref{fig:sensitivity_analysis_rate_of_change_threshold_PhysioNet}, the analysis of the plots for ROC-AUC, PR-AUC, and min(Re, Pr) metrics across different Rate of Change Threshold values (0.001, 0.01, 0.1) reveals that a threshold of 0.01 yields the best overall performance. ROC-AUC and PR-AUC both peak at 0.01, indicating optimal discriminative ability and balance between precision and recall at this threshold. While there is some variability, especially in PR-AUC and min(Re, Pr) as shown by larger error bars, the overall trends suggest that 0.01 is the most promising threshold. This threshold value offers a stable and balanced model performance, making it a robust choice for enhancing predictive capabilities.

Fig.~\ref{fig:sensitivity_analysis_insignificant_event_probability_PhysioNet} illustrates that increasing the mask probability for insignificant events generally enhances model performance.  A mask probability around 0.7 to 0.9 can offer a good trade-off, potentially leading to a better overall model performance with an acceptable level of variability. This highlights the necessity of balancing the masking of significant and insignificant events to maintain a robust and effective model.

\section{Conclusion}
\label{sec:conclusion}
We proposed a novel pretraining framework, EMIT, for irregular time series data, addressing the challenges of asynchrony, variable data density, and scarce labelled data. Our method leverages the rate of change to identify and mask significant events, enabling the model to focus on reconstructing these key points and thus learning robust, generalized features. By masking embeddings rather than raw data, EMIT ensures resilient feature learning applicable to any encoder handling time, features, and values. Experiments on Mimic-III and PhysioNet Challenge datasets demonstrate that EMIT enhances representation quality, facilitating effective fine-tuning with limited labelled data. Our findings underscore the necessity of balancing the masking of significant and insignificant events for optimal performance. This work advances irregular time series modeling and sets the stage for future research in this domain. Future work can include enhancing EMIT to model very long time series.

\section*{Acknowledgment}

This work is supported by projects IC200100022, CE200100025 funded by Australian Research Council and project AICID000001 by Australian Academy of Technological Sciences \& Engineering.

\bibliographystyle{IEEEtranS}
\bibliography{IEEEabrv,references}

% Generated by IEEEtranS.bst, version: 1.14 (2015/08/26)
\begin{thebibliography}{10}
\providecommand{\url}[1]{#1}
\csname url@samestyle\endcsname
\providecommand{\newblock}{\relax}
\providecommand{\bibinfo}[2]{#2}
\providecommand{\BIBentrySTDinterwordspacing}{\spaceskip=0pt\relax}
\providecommand{\BIBentryALTinterwordstretchfactor}{4}
\providecommand{\BIBentryALTinterwordspacing}{\spaceskip=\fontdimen2\font plus
\BIBentryALTinterwordstretchfactor\fontdimen3\font minus \fontdimen4\font\relax}
\providecommand{\BIBforeignlanguage}[2]{{%
\expandafter\ifx\csname l@#1\endcsname\relax
\typeout{** WARNING: IEEEtranS.bst: No hyphenation pattern has been}%
\typeout{** loaded for the language `#1'. Using the pattern for}%
\typeout{** the default language instead.}%
\else
\language=\csname l@#1\endcsname
\fi
#2}}
\providecommand{\BIBdecl}{\relax}
\BIBdecl

\bibitem{TCN}
S.~Bai, J.~Z. Kolter, and V.~Koltun, ``An empirical evaluation of generic convolutional and recurrent networks for sequence modeling,'' \emph{arXiv preprint arXiv:1803.01271}, 2018.

\bibitem{T-LSTM}
I.~M. Baytas, C.~Xiao, X.~Zhang, F.~Wang, A.~K. Jain, and J.~Zhou, ``Patient subtyping via time-aware lstm networks,'' in \emph{SIGKDD}, 2017.

\bibitem{SWaV}
M.~Caron, I.~Misra, J.~Mairal, P.~Goyal, P.~Bojanowski, and A.~Joulin, ``Unsupervised learning of visual features by contrasting cluster assignments,'' \emph{Advances in neural information processing systems}, 2020.

\bibitem{GRU-D}
Z.~Che, S.~Purushotham, K.~Cho, D.~Sontag, and Y.~Liu, ``Recurrent neural networks for multivariate time series with missing values,'' \emph{Scientific reports}, 2018.

\bibitem{simclr}
T.~Chen, S.~Kornblith, M.~Norouzi, and G.~Hinton, ``A simple framework for contrastive learning of visual representations,'' in \emph{International conference on machine learning}, 2020.

\bibitem{chenDynamicIllnessSeverity2018}
W.~Chen, S.~Wang, G.~Long, L.~Yao, Q.~Z. Sheng, and X.~Li, ``Dynamic {{Illness Severity Prediction}} via {{Multi-task RNNs}} for {{Intensive Care Unit}},'' in \emph{ICDM}, 2018.

\bibitem{contiformer}
Y.~Chen, K.~Ren, Y.~Wang, Y.~Fang, W.~Sun, and D.~Li, ``Contiformer: Continuous-time transformer for irregular time series modeling,'' in \emph{NeurIPS}, 2023.

\bibitem{primenet}
R.~R. Chowdhury, J.~Li, X.~Zhang, D.~Hong, R.~K. Gupta, and J.~Shang, ``Primenet: Pre-training for irregular multivariate time series,'' in \emph{AAAI}, 2023.

\bibitem{tarnet}
R.~R. Chowdhury, X.~Zhang, J.~Shang, R.~K. Gupta, and D.~Hong, ``Tarnet: Task-aware reconstruction for time-series transformer,'' in \emph{SIGKDD}, 2022.

\bibitem{GRU}
J.~Chung, C.~Gulcehre, K.~Cho, and Y.~Bengio, ``Empirical evaluation of gated recurrent neural networks on sequence modeling,'' \emph{arXiv preprint arXiv:1412.3555}, 2014.

\bibitem{imagenet}
J.~Deng, W.~Dong, R.~Socher, L.-J. Li, K.~Li, and L.~{Fei-Fei}, ``{{ImageNet}}: {{A}} large-scale hierarchical image database,'' in \emph{CVPR}, 2009.

\bibitem{BERT}
J.~Devlin, M.~Chang, K.~Lee, and K.~Toutanova, ``{BERT:} pre-training of deep bidirectional transformers for language understanding,'' in \emph{NAACL}, 2019.

\bibitem{simmtm}
J.~Dong, H.~Wu, H.~Zhang, L.~Zhang, J.~Wang, and M.~Long, ``Simmtm: {A} simple pre-training framework for masked time-series modeling,'' in \emph{NeurIPS}, 2023.

\bibitem{TS-TCC}
E.~Eldele, M.~Ragab, Z.~Chen, M.~Wu, C.~K. Kwoh, X.~Li, and C.~Guan, ``Time-{{Series Representation Learning}} via {{Temporal}} and {{Contextual Contrasting}},'' in \emph{IJCAI}, 2021.

\bibitem{triplet_loss}
J.~Franceschi, A.~Dieuleveut, and M.~Jaggi, ``Unsupervised scalable representation learning for multivariate time series,'' in \emph{NeurIPS}, 2019.

\bibitem{physionet_2012}
A.~L. Goldberger, L.~A. Amaral, L.~Glass, J.~M. Hausdorff, P.~C. Ivanov, R.~G. Mark, J.~E. Mietus, G.~B. Moody, C.-K. Peng, and H.~E. Stanley, ``Physiobank, physiotoolkit, and physionet: components of a new research resource for complex physiologic signals,'' \emph{circulation}, 2000.

\bibitem{BYOL}
J.-B. Grill, F.~Strub, F.~Altch{\'e}, C.~Tallec, P.~Richemond, E.~Buchatskaya, C.~Doersch, B.~Avila~Pires, Z.~Guo, M.~Gheshlaghi~Azar \emph{et~al.}, ``Bootstrap your own latent-a new approach to self-supervised learning,'' \emph{Advances in neural information processing systems}, 2020.

\bibitem{harutyunyan2019multitask}
H.~Harutyunyan, H.~Khachatrian, D.~C. Kale, G.~Ver~Steeg, and A.~Galstyan, ``Multitask learning and benchmarking with clinical time series data,'' \emph{Scientific data}, 2019.

\bibitem{MAE}
K.~He, X.~Chen, S.~Xie, Y.~Li, P.~Doll{\'a}r, and R.~Girshick, ``Masked autoencoders are scalable vision learners,'' in \emph{Proceedings of the IEEE/CVF conference on computer vision and pattern recognition}, 2022, pp. 16\,000--16\,009.

\bibitem{moco}
K.~He, H.~Fan, Y.~Wu, S.~Xie, and R.~Girshick, ``Momentum contrast for unsupervised visual representation learning,'' in \emph{Proceedings of the IEEE/CVF conference on computer vision and pattern recognition}, 2020.

\bibitem{seft}
M.~Horn, M.~Moor, C.~Bock, B.~Rieck, and K.~Borgwardt, ``Set functions for time series,'' in \emph{ICML}, 2020.

\bibitem{CheXpert}
J.~Irvin, P.~Rajpurkar, M.~Ko, Y.~Yu, S.~Ciurea{-}Ilcus, C.~Chute, H.~Marklund, B.~Haghgoo, R.~L. Ball, K.~S. Shpanskaya, J.~Seekins, D.~A. Mong, S.~S. Halabi, J.~K. Sandberg, R.~Jones, D.~B. Larson, C.~P. Langlotz, B.~N. Patel, M.~P. Lungren, and A.~Y. Ng, ``Chexpert: {A} large chest radiograph dataset with uncertainty labels and expert comparison,'' in \emph{AAAI}, 2019.

\bibitem{mimic_iii}
A.~E. Johnson, T.~J. Pollard, L.~Shen, L.-w.~H. Lehman, M.~Feng, M.~Ghassemi, B.~Moody, P.~Szolovits, L.~Anthony~Celi, and R.~G. Mark, ``Mimic-iii, a freely accessible critical care database,'' \emph{Scientific data}, 2016.

\bibitem{adam_opt}
D.~P. Kingma and J.~Ba, ``Adam: A method for stochastic optimization,'' \emph{arXiv preprint arXiv:1412.6980}, 2014.

\bibitem{lipton2016directly}
Z.~C. Lipton, D.~Kale, and R.~Wetzel, ``Directly modeling missing data in sequences with rnns: Improved classification of clinical time series,'' in \emph{MLHC}, 2016.

\bibitem{lipton2015learning}
Z.~C. Lipton, D.~C. Kale, C.~Elkan, and R.~C. Wetzel, ``Learning to diagnose with {LSTM} recurrent neural networks,'' in \emph{ICLR}, 2016.

\bibitem{focal}
S.~Liu, T.~Kimura, D.~Liu, R.~Wang, J.~Li, S.~Diggavi, M.~Srivastava, and T.~Abdelzaher, ``Focal: Contrastive learning for multimodal time-series sensing signals in factorized orthogonal latent space,'' in \emph{NeurIPS}, 2024.

\bibitem{roberta}
Y.~Liu, M.~Ott, N.~Goyal, J.~Du, M.~Joshi, D.~Chen, O.~Levy, M.~Lewis, L.~Zettlemoyer, and V.~Stoyanov, ``Roberta: A robustly optimized bert pretraining approach,'' \emph{arXiv preprint arXiv:1907.11692}, 2019.

\bibitem{infots}
D.~Luo, W.~Cheng, Y.~Wang, D.~Xu, J.~Ni, W.~Yu, X.~Zhang, Y.~Liu, Y.~Chen, H.~Chen, and X.~Zhang, ``Time series contrastive learning with information-aware augmentations,'' in \emph{AAAI}, 2023.

\bibitem{PTM_TS_Review}
Q.~Ma, Z.~Liu, Z.~Zheng, Z.~Huang, S.~Zhu, Z.~Yu, and J.~T. Kwok, ``A survey on time-series pre-trained models,'' \emph{arXiv preprint arXiv:2305.10716}, 2023.

\bibitem{timenet}
P.~Malhotra, V.~TV, L.~Vig, P.~Agarwal, and G.~Shroff, ``Timenet: Pre-trained deep recurrent neural network for time series classification,'' in \emph{ESANN}, 2017.

\bibitem{MHCCL}
Q.~Meng, H.~Qian, Y.~Liu, L.~Cui, Y.~Xu, and Z.~Shen, ``{MHCCL:} masked hierarchical cluster-wise contrastive learning for multivariate time series,'' in \emph{AAAI}, 2023.

\bibitem{basisformer}
Z.~Ni, H.~Yu, S.~Liu, J.~Li, and W.~Lin, ``Basisformer: Attention-based time series forecasting with learnable and interpretable basis,'' in \emph{NeurIPS}, 2024.

\bibitem{peng2021sequential}
X.~Peng, G.~Long, T.~Shen, S.~Wang, and J.~Jiang, ``Sequential diagnosis prediction with transformer and ontological representation,'' in \emph{2021 IEEE International Conference on Data Mining (ICDM)}, 2021.

\bibitem{peng2019temporal}
X.~Peng, G.~Long, T.~Shen, S.~Wang, J.~Jiang, and M.~Blumenstein, ``Temporal self-attention network for medical concept embedding,'' in \emph{2019 IEEE international conference on data mining (ICDM)}, 2019.

\bibitem{GPT}
A.~Radford, K.~Narasimhan, T.~Salimans, I.~Sutskever \emph{et~al.}, ``Improving language understanding by generative pre-training,'' 2018.

\bibitem{multimodal_clincial_time_series_cl}
A.~Raghu, P.~Chandak, R.~Alam, J.~Guttag, and C.~Stultz, ``Contrastive pre-training for multimodal medical time series,'' in \emph{NeurIPS Workshop}, 2022.

\bibitem{Rubanova2019LatentOD}
Y.~Rubanova, T.~Q. Chen, and D.~K. Duvenaud, ``Latent ordinary differential equations for irregularly-sampled time series,'' in \emph{NeurIPS}, 2019.

\bibitem{shukla2021deep}
S.~N. Shukla, ``Deep learning models for irregularly sampled and incomplete time series,'' 2021.

\bibitem{IP-NET}
S.~N. Shukla and B.~M. Marlin, ``Interpolation-prediction networks for irregularly sampled time series,'' \emph{arXiv preprint arXiv:1909.07782}, 2019.

\bibitem{sand}
H.~Song, D.~Rajan, J.~Thiagarajan, and A.~Spanias, ``Attend and diagnose: Clinical time series analysis using attention models,'' in \emph{AAAI}, 2018.

\bibitem{dropout}
N.~Srivastava, G.~Hinton, A.~Krizhevsky, I.~Sutskever, and R.~Salakhutdinov, ``Dropout: a simple way to prevent neural networks from overfitting,'' \emph{JMLR}, 2014.

\bibitem{strats}
S.~Tipirneni and C.~K. Reddy, ``Self-supervised transformer for sparse and irregularly sampled multivariate clinical time-series,'' \emph{ACM TKDD}, 2022.

\bibitem{TNC}
S.~Tonekaboni, D.~Eytan, and A.~Goldenberg, ``Unsupervised representation learning for time series with temporal neighborhood coding,'' in \emph{ICLR}, 2021.

\bibitem{transformer}
A.~Vaswani, N.~Shazeer, N.~Parmar, J.~Uszkoreit, L.~Jones, A.~N. Gomez, {\L}.~Kaiser, and I.~Polosukhin, ``Attention is all you need,'' in \emph{NeurIPS}, 2017.

\bibitem{contrast_everything}
Y.~Wang, Y.~Han, H.~Wang, and X.~Zhang, ``Contrast everything: {A} hierarchical contrastive framework for medical time-series,'' in \emph{NeurIPS}, 2023.

\bibitem{btsf}
L.~Yang and S.~Hong, ``Unsupervised time-series representation learning with iterative bilinear temporal-spectral fusion,'' in \emph{ICML}, 2022.

\bibitem{TST}
G.~Zerveas, S.~Jayaraman, D.~Patel, A.~Bhamidipaty, and C.~Eickhoff, ``A transformer-based framework for multivariate time series representation learning,'' in \emph{SIGKDD}, 2021.

\bibitem{timeseries_SSL_survey}
K.~Zhang, Q.~Wen, C.~Zhang, R.~Cai, M.~Jin, Y.~Liu, J.~Y. Zhang, Y.~Liang, G.~Pang, D.~Song, and S.~Pan, ``Self-supervised learning for time series analysis: Taxonomy, progress, and prospects,'' \emph{IEEE TPAMI}, 2024.

\bibitem{TF-C}
X.~Zhang, Z.~Zhao, T.~Tsiligkaridis, and M.~Zitnik, ``Self-supervised contrastive pre-training for time series via time-frequency consistency,'' in \emph{NeurIPS}, 2022.

\end{thebibliography}

\end{document}